\pdfoutput=1
\documentclass[11pt]{article}
\usepackage[utf8]{inputenc}
\usepackage[T1]{fontenc}

\usepackage[final]{acl}

\usepackage{times}
\usepackage{latexsym}
\usepackage{enumitem}
\usepackage{booktabs}

\definecolor{mygray}{HTML}{555555}
\definecolor{mygreen}{HTML}{b4e081}
\definecolor{myred}{HTML}{f4cccc}
\definecolor{reallygreen}{HTML}{56724a}
\definecolor{reallyred}{HTML}{a95544}

\usepackage[T1]{fontenc}
\usepackage[utf8]{inputenc}

\usepackage{microtype}

\usepackage{inconsolata}
\usepackage{tikz}
\usetikzlibrary{shapes.geometric, arrows, positioning, fit}

\usepackage{amsmath}
\usetikzlibrary{positioning, shapes, shadows, arrows.meta}

\usepackage{graphicx}
\usepackage{url}
\usepackage{xspace}

\title{MultiHoax: A Dataset of Multi-hop False-Premise Questions}

\author{Mohammadamin Shafiei\thanks{\enspace Equal contribution.}\textsuperscript{1}, \space 
  Hamidreza Saffari$^{*}$\textsuperscript{2}, \space  
  Nafise Sadat Moosavi\textsuperscript{3} \\
\textsuperscript{1}University of Milan, 
\textsuperscript{2}Politecnico di Milano, 
\textsuperscript{3}University of Sheffield \\
\texttt{m.shafieiapoorvari@studenti.unimi.it} \\
\texttt{hamidreza.saffari@mail.polimi.it} \\
\texttt{n.s.moosavi@sheffield.ac.uk} \\
}
\begin{document}
\maketitle
\begin{abstract}
As Large Language Models are increasingly deployed in high-stakes domains, their ability to detect false assumptions and reason critically is crucial for ensuring reliable outputs. False-premise questions (FPQs) serve as an important evaluation method by exposing cases where flawed assumptions lead to incorrect responses. While existing benchmarks focus on single-hop FPQs, real-world reasoning often requires multi-hop inference, where models must verify consistency across multiple reasoning steps rather than relying on surface-level cues.
To address this gap, we introduce MultiHoax, a benchmark for evaluating LLMs' ability to handle false premises in complex, multi-step reasoning tasks. Our dataset spans seven countries and ten diverse knowledge categories, using Wikipedia as the primary knowledge source to enable factual reasoning across regions.
Experiments reveal that state-of-the-art LLMs struggle to detect false premises across different countries, knowledge categories, and multi-hop reasoning types, highlighting the need for improved false premise detection and more robust multi-hop reasoning capabilities in LLMs.\footnote{The MultiHoax dataset is publicly available at \url{https://github.com/Mamin78/MHFPQ}}

%As Large Language Models(LLMs) become more widely used in high-stakes domains, their ability to reason critically and detect flawed assumptions is essential for ensuring reliable outputs. False-premise questions (FPQs) play a crucial role in evaluating AI models, as they expose scenarios where incorrect assumptions can lead to misleading or inaccurate responses. 
%Existing benchmarks primarily assess single-hop FPQs, where false premises can be identified in a single reasoning step. In contrast, real-world reasoning often involves multi-hop inference, requiring models to connect multiple facts and verify consistency across steps. Multi-hop questions introduce an additional layer of complexity by requiring models to decompose complex queries into sequential inferential steps rather than relying on surface-level pattern recognition.
%To address these challenges, we introduce Multi-Hop False-Premise Questions (MHFPQs), a novel benchmark where false premises are implicitly embedded within multi-hop reasoning tasks, alongside plausible distractors and a correct choice. To enable comprehensive evaluation, we compiled a cross-country dataset spanning 7 countries with varying levels of resource availability and 10 categories per country, using Wikipedia as the primary knowledge source.
%Our experiments reveal that current state-of-the-art LLMs not only fail to detect false information in MHFPQs in general but also show varying accuracies across different countries and categories. These findings underscore the need for improved uncertainty handling.
\end{abstract}

\section{Introduction}

% \begin{figure}[!t]
%   \centering
%   \includegraphics[width=\columnwidth]{figures/mhfpqs_resized copy.png}
%   \caption{A sample MHFPQ from the history category related to Iran.}
%   \label{fig:example}
% \end{figure}

\begin{figure}[t]
  \centering
\begin{tikzpicture}[
  every node/.style={font=\small},
  % label/.style={font=\scriptsize, text width=1.2cm, fill=mygray, inner sep=5pt, text=white},
  label/.style={font=\scriptsize,  text width=1.2cm, inner sep=5pt},
  greenbox/.style={draw=reallygreen, font=\scriptsize,  fill=mygreen, inner sep=5pt},
  redbox/.style={draw=reallyred, font=\scriptsize,  fill=myred, inner sep=5pt},
  bluebox/.style={draw=black, fill=blue!20, rounded corners, inner sep=3pt},
  highlight/.style={font=\scriptsize, text width=15cm, rectangle, draw=none, rounded corners=2pt},
  arrow/.style={->, thick},
  x=0.5cm, y=0.6cm
]

% Main question block
\node[highlight, align=left, text width=16cm] (q) at (0,0) {
  \textbf{\footnotesize Which \textcolor{red}{Iranian wrestler won gold in the Men's freestyle \\ 125 kg} at \textcolor{olive}{the first Olympics when Zahra Nemati was \\the flag bearer}?}
};

% Options
\node[label, below=1.4 of q.west, anchor=west] (a1) {Komeil Ghasemi};
\node[label, right=0.65 of a1] (a2) {Ghasem Rezaei};
\node[label, right=0.65 of a2] (a3) {Hassan Yazdani};
\node[label, right=0.65 of a3, fill=mygray, text=white, text width=1.5cm] (a4) {I do not know};

% Step 1
\node[greenbox, below=3.4 of q.west, anchor=west, text width=7.35cm] (s1) {\footnotesize
\textbf{1.} \textbf{The first hop:} \\
\vspace{\baselineskip}
At which Olympics did Zahra Nemati carry the flag for the first time?
};
% \node[right=0.2 of s1, align=left] (d1) {\scriptsize Islamic Azad University (IAU)};

\node[below=2.2 of s1.west, anchor=west, text width=7.35cm] (e1) { \textbf{The answer:} Zahra Nemati carried the flag for the first time at the \textbf{2016 Rio Olympics}.};

% Step 2
\node[redbox, below=2.5 of e1.west, anchor=west, text width=7.35cm] (s2) {\footnotesize
\textbf{2.}  \textbf{The second hop:} \\ 
\vspace{\baselineskip}
Which Iranian wrestler won gold in the Men's freestyle 125 kg at the 2016 Rio Olympics?
};
% \node[right=0.2 of s2, text=black] (d2) {\scriptsize IAU was founded in \textbf{1982}, so it could not have any students in \textbf{1980}.};

\node[below=3 of s2.west, anchor=west, text width=7.35cm] (e2) { \textbf{Falsehood:} \textit{No Iranian wrestler won gold in the Men's freestyle 125 kg at the 2016 Rio Olympics.} The only Iranian wrestler to win gold in the Men's freestyle at the 2016 Rio Olympics was Hassan Yazdani in \textbf{74 kg}.};
\end{tikzpicture}
  \caption{A sample MHFPQ from the Sports category related to Iran.}
  \label{fig:example}
\end{figure}

In recent years, the evolution of large language models (LLMs) has demonstrated their immense potential across a wide range of natural language processing tasks. However, despite their impressive successes in many domains, their ability to recognize and properly handle false premises in reasoning tasks remains a significant challenge \cite{hu-etal-2023-wont, yu-etal-2023-crepe}. When engaging with false premises, an LLM should be able to identify and reject flawed assumptions rather than proceed as if they were valid.
%there remains a critical gap in their abilities, especially the ability to identify and handle false information.

False premises can appear in various forms, such as misleading statements, logically inconsistent claims, or factually incorrect contextual narratives \cite{leite2023detecting, NEURIPS2023_9cb2a749, ghosh2024logical, galitsky2024truth, chen2023can, yamin2024failure, zhang2024revealing, csonge2015moving}. A common approach to evaluating an LLM’s ability to handle false premises is through question-answering (QA) tasks, where a question implicitly contains a misleading or incorrect assumption \cite{yu-etal-2023-crepe, hu-etal-2023-wont, kim-etal-2023-qa}. For example, a well-performing LLM should recognize that \textit{``Which year did Albert Einstein win the Nobel Prize in Chemistry?''} contains a false premise—Einstein never won a Nobel Prize in Chemistry—and responds appropriately, rather than attempting to provide an incorrect or misleading answer.

Existing benchmarks focus on single-hop false premise detection, where the incorrect assumption can be identified within a single reasoning step \cite{yu-etal-2023-crepe, hu-etal-2023-wont, kim-etal-2023-qa}. However, multi-hop reasoning presents a greater challenge, as it requires models to connect multiple pieces of information across multiple inference steps before arriving at an answer \cite{mavi2022survey}. Unlike single-hop questions, multi-hop questions (MHQs) require models to derive intermediate conclusions, making them a critical area of research in the evaluation and improvement of LLM reasoning abilities \cite{mavi2022survey, chen2024llm, yang2024large, tang2024multihop, chakraborty2024multi}. The complexity of MHQs arises from the necessity of bridging multiple facts, understanding implicit dependencies between reasoning steps, and maintaining logical consistency throughout the inference process \cite{mavi2022survey}.

While MHQs are already challenging, the problem becomes even more difficult when false premises are embedded within the reasoning chain, requiring models not only to answer questions correctly but also to detect and reject incorrect assumptions at intermediate steps. To evaluate LLMs' ability to handle this challenge, we introduce a novel benchmark of Multi-Hop False-Premise Questions (MHFPQs), called \textbf{\texttt{MultiHoax}}, that combines the difficulty of false premise questions and multi-hop question answering. MHFPQs are designed to test whether LLMs can detect false premises that appear at one or more reasoning steps rather than simply providing an answer based on a flawed assumption.

Figure~\ref{fig:example} shows an example of an MHFPQ from the sports category related to Iran. Answering this question requires a structured reasoning process. First, the model must determine the edition of the Olympics where Zahra Nemati served as Iran's flag bearer for the first time, namely, the 2016 Rio Olympics. Second, it must identify the Iranian wrestler who won the gold medal in the Men’s freestyle 125 kg category at those Olympics. This introduces a false premise: no Iranian wrestler won gold in the 125 kg category at the 2016 Olympics. The only Iranian gold medalist in Men’s freestyle wrestling at those Games was Hassan Yazdani, who competed in the 74 kg weight class. %First, the model must correctly identify the first time when Zahra Nemati was the flag carrier at the Olympics. In the second reasoning step, the model must provide the name of the Iranian wrestler who won the Men's freestyle gold when Zahra Nemati was the flag carrier at the Olympics for the first time. This is a false question as Zahra Nemati carried the flag at the 2016 Rio Olympics for the first time, at which no Iranian wrestler won the freestyle 125 kg gold.

To ensure realism and difficulty, we carefully designed the set of possible answer choices in the MHFPQ dataset. Each question includes believable distractors, as possible, that align with historical facts, making the task particularly challenging for LLMs. For instance, in the example from Figure~\ref{fig:example}, the distractors are real Iranian wrestlers. Moreover, Ghassem Rezaei and Komeil Ghassemi both won gold in freestyle wrestling at the 2012 London Olympics. Hassan Yazdani also won the gold at the 2016 Olympics, but was not competing in the 125 kg division. This ensures that incorrect answers remain plausible while still being contextually invalid. Without detecting the false premise, LLMs may be misled into selecting one of the plausible but incorrect answers, emphasizing the necessity for models to engage in deep contextual reasoning rather than relying on surface-level fact retrieval.

The dataset spans 7 countries and 10 distinct categories, ensuring broad diversity across historical, cultural, and geopolitical contexts. To generate the MHFPQs, we extracted relevant information from Wikipedia pages with the assistance of the Claude model \footnote{\url{https://www.anthropic.com/}}. Each question is paired with three closely related distractors and a single ``I do not know'' option, which serves as the only valid response in the presence of a false premise. By structuring the benchmark in this way, we aim to assess LLMs’ ability to identify and reject incorrect assumptions embedded within multi-hop reasoning chains, rather than merely selecting the most superficially plausible response.

We conduct a comprehensive evaluation of six open-source and closed-source models using our \texttt{MultiHoax} dataset of 700 carefully reviewed questions. Our results reveal suboptimal performance across all models, categories, and countries, indicating that current systems struggle with detecting falsehoods in multi-hop reasoning tasks. This resource enables rigorous analysis of multi-hop reasoning under false premises and opens new research directions in question answering that require rejection of complex, embedded assumptions.

\section{Related Work}
\paragraph{False Information and False Premises.} 
False information is a broad and multifaceted issue, encompassing explicitly misleading claims from social networks \cite{ma-etal-2022-open-topic, rode-hasinger-etal-2022-true}, blogs \cite{okazaki-etal-2013-extracting}, news sources \cite{long-etal-2017-fake, qiao-etal-2020-language, huang-etal-2023-faking}, and online forums \cite{yu-etal-2023-crepe}. However, false premises are a distinct challenge since they are not always deliberate misinformation but rather incorrect assumptions that lead to flawed logical conclusions. These premises span various domains, including historical events \cite{gabba1981true, smith2001false}, religion \cite{kutlu2022analysis}, and sports \cite{dimov2021recognition}, making them particularly complex to detect.

\paragraph {False-Premise Questions.} 
FPQs assess a model's ability to reason over implicit false assumptions about properties, actions, scope, existence, events, logic, or causality \cite{hu-etal-2023-wont, yu-etal-2023-crepe}. For example, ``How many eyes does the sun have?'' wrongly assumes the sun has eyes \cite{hu-etal-2023-wont}.
Since FPQs require models to identify and reject flawed premises rather than simply answering the question, LLMs often fail to recognize implicit falsehoods, leading to misleading or nonsensical responses \cite{yu-etal-2023-crepe}. This inability to reject false premises poses risks for misinformation propagation and model alignment with human expectations.

Early studies on FPQs have focused on obvious falsehoods that humans can easily detect. For instance, \citet{hu-etal-2023-wont} introduced a dataset of simple FPQs, such as ``What is the most common color of human’s wings?''. While earlier LLMs struggled with these, recent versions of the LLMs handle them easily, raising the need to evaluate models on more subtle false premises. Similarly, \citet{kim-etal-2023-qa} proposed a dataset evaluating models on questions containing false or unverifiable assumptions. However, their dataset includes unverifiable claims—statements that may become true over time. For example, ``When is Steven Universe Season 5 coming to Hulu?'' assumes the event has not yet occurred in order to be false, but this assumption could later become valid. This distinction makes their dataset less suitable for assessing persistent false premises.
Another line of research has focused on false-presupposition questions extracted from online forums. \citet{yu-etal-2023-crepe} introduced a dataset of FPQs sourced from Reddit, such as ``How exactly is current stored in power plants?''—a misleading assumption since electric current is not stored. However, their dataset is primarily limited to scientific and technical topics, lacking the broad factual diversity required for a general evaluation of false premises.

Beyond FPQs, unanswerable questions have also been explored in related research. \citet{zhao-etal-2024-couldve} focus on document-grounded unanswerable questions, where a question lacks supporting information within a given document. Their contribution is in evaluating models’ ability to reformulate unanswerable questions into answerable ones, rather than directly assessing how LLMs handle false premises in an open-domain setting. Additionally, \citet{lin-etal-2022-truthfulqa} study truthful question answering, focusing on questions that elicit misconceptions from humans. Relatedly, \citet{zhang-etal-2024-r} and \citet{peng2024scopeqa} examine how LLMs handle questions beyond their knowledge, where the correct response should be ``I do not know''. These studies, however, focus on uncertain or conflicting information rather than inherently false premises.

\paragraph{Multi-Hop False-Premise Reasoning.}

Most FPQ benchmarks focus on single-hop false premises, yet real-world misinformation often involves multi-step reasoning, making it significantly harder to detect. Multi-hop reasoning questions challenge LLMs by requiring them to connect multiple facts across inference steps. While extensive research has explored general MHQ understanding in LLMs \cite{yang-etal-2018-hotpotqa, rajabzadeh2023multimodal, park2023graph, chen2024llm}, MHQs embedding false premises remain largely unstudied. Existing benchmarks assess multi-hop factual reasoning but do not evaluate whether LLMs can detect and reject false premises within reasoning chains. This gap is critical, as misinformation is rarely an isolated error—false premises are often interwoven across reasoning steps, making them subtle, plausible, and difficult to refute. Evaluating how LLMs handle multi-hop false premises is essential for enhancing their robustness against misinformation and ensuring logical consistency in complex reasoning tasks.

\citet{daswani2024syn} attempt to address this with adversarial multi-hop false premise questions, modifying HotpotQA \cite{yang-etal-2018-hotpotqa}. Their approach replaces the title of a supporting document with a similar but unrelated distractor, selected based on shared Wikipedia categories. For example, Roger O. Egeberg and Steven K. Galson both fall under American public health doctors, making them interchangeable under this method. However, this technique relies on structured entity swaps rather than embedding implicit falsehoods, making it unclear whether a question contains a truly false premise or is merely unverifiable. Additionally, this approach can lead to unnatural question phrasing, limiting its applicability in assessing real-world false premise detection.

Overall, LLMs struggle with imperfect information, conflicting evidence, and questions beyond their knowledge scope \cite{lin-etal-2022-truthfulqa, zhang-etal-2024-r, peng2024scopeqa, kazemi2024boardgameqa, payandeh-etal-2024-susceptible, shaier-etal-2024-adaptive, park-lee-2024-toward, longpre-etal-2021-entity, chen-etal-2022-rich}. They are also susceptible to distraction by irrelevant details, including false premises, often leading to incorrect, misleading, or fabricated responses \cite{park-lee-2024-toward, yu-etal-2023-crepe, kim-etal-2023-qa, hu-etal-2023-wont, asai-choi-2021-challenges}. These findings underscore a critical gap in evaluating multi-step false premise reasoning, as prior benchmarks focus on single-hop FPQs and adversarial perturbations, failing to capture the complexity of false premises embedded within structured factual reasoning. To address this, we introduce {\texttt{MultiHoax}}, a new benchmark of multi-hop FPQs, designed to assess LLMs' ability to detect and reject false premises within reasoning chains. By including implicit false factual claims across diverse country-related topics, our benchmark enables a more nuanced evaluation of LLMs’ ability to detect and reject false premises embedded in complex, multi-hop reasoning chains.

\paragraph{Multi-Regional and Multi-Cultural Resources.}
Our dataset aligns with multi-cultural and multi-regional NLP research, which examines how LLMs process knowledge across different geographic and cultural contexts. While much prior work has focused on subjective tasks like norms and values \cite{ziems2023normbank, cheng2023compost, fung2024massively, shi2024culturebank, han2023reading, saffari-etal-2025-introduce}, recent studies show that multicultural NLP also includes objective knowledge, such as region-specific facts, historical events, and socio-political contexts \cite{koto-etal-2024-arabicmmlu, li-etal-2024-cmmlu, koto-etal-2023-large, son2024kmmlu, kim-etal-2024-click}. This highlights the need to evaluate how LLMs handle factual reasoning across diverse regions, particularly in multi-hop tasks involving false premises.

Despite growing interest in multi-regional fact verification, most work focuses on binary misinformation detection rather than multi-hop false premise reasoning \cite{thaher2021intelligent, sabzali2022fake, sheng2022characterizing}. A global benchmark for multi-hop false premise detection remains missing, even though factual inaccuracies vary significantly across regions. 
\texttt{MultiHoax} addresses this gap by introducing diverse, context-rich questions across seven countries and ten knowledge domains, enabling evaluation of LLMs' reasoning over embedded falsehoods in culturally grounded settings.

\section{MultiHoax Dataset}
This section describes the proposed dataset, designed to evaluate multi-hop false premise reasoning across diverse countries, categories, and question types. The dataset structure enables a systematic analysis of how LLMs handle false premises within complex reasoning tasks.

\subsection{Dataset Framework}
Each \textit{MultiHoax}'s instance consists of the following components:

\begin{itemize}[noitemsep, topsep=2pt]
    \item \textbf{Question and Answer Choices:} Each question contains at least a false premise and is paired with four answer choices, three plausible distractors, and one ``I do not know''. The latter is positioned randomly to prevent models from exploiting positional bias.
    
     \item \textbf{False Premise Explanation:} A brief description clarifies why the assumption in the question is incorrect.
    
    \item \textbf{Country and Category:} The dataset spans seven countries (China, France, Germany, Iran, Italy, the United Kingdom, and the United States) and is categorized into ten knowledge domains, including food, sports, geography, education, history, entertainment, religion, science \& technology, arts \& literature, and holidays \& leisure. 
    
    \item \textbf{Types of False Premises and Answer Types:} Each question is annotated with the type of false premise it contains, following the taxonomy introduced by \newcite{hu-etal-2023-wont}. The five main types are Property, Event, Entity, and Scope. For example, the question \textit{``Which award did Venkatraman Ramakrishnan receive first: the Shaw Prize in Life Science and Medicine or the Lasker Award?''} falls under the event type, since it implies an event that never happened because he never won the mentioned awards.\footnote{Table~\ref{tab:fptypescount} in the appendix presents the distribution of false premise types in the dataset, along with definitions of each type.} 
    %Inspired by \citet{hu-etal-2023-wont}, we annotate each question based on how it includes a false premise. The primary types of false premises include Property, Event, Entity, Scope, and Index. For instance, the question \textit{``Which award did Venkatraman Ramakrishnan receive first: the Shaw Prize in Life Science and Medicine or the Lasker Award?''} falls under the Event type, as it falsely implies an event, that is winning these awards by him, while in reality, he did not.
   Furthermore, following \citet{yang-etal-2018-hotpotqa}, we classify answer types such as person, location, event, number, and common nouns. The answer type shows what the question is asking for.\footnote{Table~\ref{tab:optcount} in the appendix provides a detailed breakdown of answer option types across the dataset, including an example for each type.}
    
    \item \textbf{Multi-Hop Reasoning Type:} Following \citet{mavi2022survey}, we categorize why a question requires multi-step inference into five types: (1) Named Entity Reasoning: The question requires connecting two facts through an intermediate entity that links them logically, (2) Temporal Reasoning: An intermediate step involves identifying a specific time reference to answer the question correctly, (3) Geographical Reasoning: The reasoning process depends on understanding locations, spatial relationships, or geographic entities, (4) Intersection Reasoning: The answer is determined by an entity that satisfies multiple overlapping conditions, and (5) Comparison Reasoning: The question requires comparing attributes, facts, or values across multiple entities to arrive at the correct conclusion.\footnote{Table \ref{tab:mhtypecount} in the appendix presents the distribution of each multi-hop type in the dataset.}
    
    \item \textbf{Wikipedia Grounding:} Each question is linked to a relevant Wikipedia page for factual grounding.
\end{itemize}

\subsection{Wikipedia Document Collection}
We developed a pipeline to extract Wikipedia pages relevant to each country and category using ChatGPT-4o's search tool. To select the set of pages, each page was evaluated based on three criteria: relevance to the category, association with the specified country, and existence. ``Existence'' ensures that the provided link leads to an actual Wikipedia page rather than just an important but undocumented topic. If a page was missing, the model was prompted to complete the list by suggesting alternatives. Ultimately, we collected 15 Wikipedia pages per country and category and extracted their content.

\subsection{Question Generation}
After collecting relevant documents, we developed a structured process for generating MHFPQs. First, we instructed Claude 3.5 Haiku to extract 15 facts per document using a fact-extraction prompt. Then, based on these facts, we prompted the model to generate MHFPQs. All prompts are detailed in Appendix~\ref{sec:appendixa}.
Due to fundamental differences between Bridge Entity-based MHQs (named, geographical, and temporal entities) and other types (intersection and comparison), we used a separate prompt for each category. We requested three questions from the former and two from the latter but did not enforce specific subtypes (e.g., one intersection, one comparison), as not all documents contained relevant supporting facts.
Additionally, we avoided first generating MHQs and then falsifying them, as this could limit the variety of false information types \cite{daswani2024syn} and might not ensure incorrectness.

\begin{table*}[ht]
\centering
\renewcommand{\arraystretch}{0.9} 
\setlength{\tabcolsep}{4pt} 
\footnotesize 
\resizebox{\textwidth}{!}{ 
\begin{tabular}{p{4.5cm} p{4.5cm} p{2.5cm} p{2.5cm} p{2.5cm}}
\toprule
\textbf{Question} & \textbf{Description} & \textbf{MH Type} & \textbf{Answer Type} & \textbf{FP Type} \\
\midrule
Which Bundesliga team, Bayern Munich or Borussia Dortmund, has the stadium with the largest seating capacity among clubs that have won the FIFA World Cup? 
& Clubs do not win the FIFA World Cup, only national teams (like Germany, Brazil, etc.) do. Spotify Camp Nou and Estadio Santiago Bernabéu have the largest capacities among clubs that have won the FIFA Club World Cup, not the FIFA World Cup. 
& Comparison & Group/Org & Event \\
% \midrule
% Which German player scored a hat-trick against Portugal and then went on to score the winning goal for Germany in the 2014 FIFA World Cup final?
% & Thomas Müller scored a hat-trick against Portugal, but Mario Götze scored the winning goal for Germany in the final against Argentina.
% & Intersection & Person & Event \\

\midrule
Who both served as as a volunteer in the Illinois Militia during the Black Hawk War seeing lots of combat during his tour, and also was among the assassinated presidents of the US?
& Abraham Lincoln served as a volunteer in the Illinois Militia April 21, 1832 – July 10, 1832, during the Black Hawk War. However, Lincoln never saw combat during his tour. He was assassinated as well.
& Intersection & Person & Property \\

\midrule
In which country, besides the U.S. and Italy, was the 1972 American epic gangster film directed by Francis Ford Coppola filmed?
& The Godfather (1972) was filmed exclusively in locations around New York City and Sicily, with no scenes shot in other countries.
& Named Entity & Location & Event \\

\bottomrule
\end{tabular}
}
\caption{Examples of MHFPQs from MultiHoax.}
\label{tab:MHFPQSEX}
\end{table*}

\iffalse
\begin{table*}[ht]
\centering
\begin{tabular}{p{4cm}p{4cm}p{2cm}p{2cm}p{2cm}}
\hline
Question & Description & MH type & Answer type & FP type \\
\hline
Which Bundesliga team, Bayern Munich or Borussia Dortmund, has the stadium with the largest seating capacity among clubs that have won the FIFA World Cup? & None of them. Spotify Camp Nou and Estadio Santiago Bernabéu have the largest capacities among clubs that have won the FIFA Club World Cup, not the FIFA World Cup. & Comparison & Group\_or\_Org & Event \\
\hline
Which École Normale Supérieure alumnus both won a Nobel Prize and was a member of the Institut de France in the 20th century? & The École Normale Supérieure has many distinguished alumni, including 14 Nobel laureates and hundreds of Institut de France members, but no one has achieved both distinctions. & Intersection & Person & Property \\
\hline
\end{tabular}
\caption{Examples of MHFPQs from our resource.}
\label{tab:MHFPQSEX}
\end{table*}
\fi

\subsection{Curated Selection and Expert Review}
After generating the questions for each category and country, an expert reviewer evaluated the generated questions and selected 10 MHFPQs for each combination of category and country. The selection was guided by a structured, multi-step approach to ensure the quality, validity, and plausibility of the questions.

The first step in the selection process was to verify whether a question adhered to the MH structure. Ensuring diversity in MH question types was a key objective. If a question was not initially formatted as an MHQ but could be converted into one, it passed this initial filter. For instance, the question ``What was the name of the Achaemenid ruler who appointed Cyrus as governor of the Median Empire?'' is inherently a single-hop reasoning question. However, by introducing an additional reasoning step, such as asking about the ruler’s son, it could be transformed into an MHQ: ``Who was the son of the Achaemenid ruler who appointed Cyrus as governor of the Median Empire?''.

Next, the question needed to contain at least one universally false piece of information—meaning the falsehood had to be global rather than merely incorrect within the context of the associated document. To ensure this criterion was met, only questions that demonstrably contained globally false statements were selected. This verification process followed a rigorous three-step approach. First, the reviewer traced each question back to its corresponding fact or set of facts. Second, these facts were cross-checked against the information found in the relevant document. Finally, the reviewer assessed whether the question was indeed incorrect relative to both the established facts and the document. If the model did not generate enough false information, the reviewer modified the question by introducing falsehoods aligned with the dataset’s predefined types of misinformation. However, if adding such falsehoods was not feasible, the question was discarded.

The third step involves explaining why the question is incorrect. To achieve this, the model's explanation is evaluated based on the previous step’s results. If it fails to address the false information, the reviewer provides a more detailed clarification.

The final step involved verifying the plausibility of the answer choices. The reviewer ensured all options were contextually relevant by referring to the corresponding section of the document. If the model’s initial options were unrelated—either due to a change in the question’s focus or the model’s poor performance in generating relevant choices—the reviewer replaced them with more appropriate alternatives.

An example of this procedure is illustrated in Figure~\ref{fig:example}. The initial question, generated by Claude 3.5 Haiku, was: \textit{At which Olympics did Zahra Nemati carry the flag for
the first time?} While this was a factual question, it was not an MHQ, but it could be transformed into one, as shown in Figure~\ref{fig:example}.

Initially, the question contained no falsehoods—it simply inquired about an event that happened in reality. To introduce false information, we modified the question to ask about an event that did not happen, while adding another reasoning hop. Asking about the Iranian freestyle wrestler who won gold in the 125 kg division at the 2016 Olympics contains a falsehood, as no Iranian wrestler did so at the 2016 Olympics.
After these steps, the question was transformed into an MHFPQ, such as: \textit{Which Iranian wrestler won gold in the Men’s freestyle 125 kg at the first Olympics when Zahra Nemati was the flag bearer?} 

To generate plausible answer choices, we included real Iranian wrestlers. Additionally, we selected the most relevant wrestlers. For example, Komeil Ghasemi won a gold medal in the men's freestyle 120 kg event at the 2012 Summer Olympics. Moreover, Ghasem Rezaei was a former Greco-Roman wrestler and an Olympic gold medalist. Additionally, the only Iranian wrestler to win gold in the Men’s freestyle at the 2016 Rio Olympics was Hassan Yazdani in 74 kg. This set of wrestlers provides three relevant and distracting options.

% An example of this procedure is illustrated in Figure \ref{fig:example}. The initial question generated by Claude 3.5 Haiku was: \textit{Who was the Prime Minister of Iran when Darius the Great's statue in Persepolis was commissioned?}. The model also provided Reza Shah Pahlavi, Mohammad Mosaddegh, and Hassan Rouhani as possible answer choices.
% However, the false premise was too obvious. As the model itself explained: \textit{Darius the Great ruled the Achaemenid Empire in the 6th-5th centuries BCE, long before the modern nation of Iran and its government structures existed. There was no Prime Minister of Iran during Darius's reign.} Due to this significant time gap, the question lacked subtle misinformation. 

% To introduce a more nuanced falsehood, the reviewer modified the question to: \textit{Who was the wife of the eldest son of Hystaspes who became the king of Iran, when his statue in Persepolis was commissioned by the order of Cyrus the Great?}. Here, the false premise—that Cyrus the Great commissioned Darius the Great’s statue—is less obvious because the time gap is reduced, and the context remains within the same historical period of ancient Iran. Additionally, to ensure the question met MH criteria, it asked for the name of Darius the Great’s wife. To provide plausible answer choices, three names were selected from his six known wives: Atossa, Artystone, Parmys, Phratagune, Phaedymia, and the daughter of Gobryas.

\begin{table*}[ht]
\centering
\renewcommand{\arraystretch}{1.2} % Adjust row height for better spacing
\setlength{\tabcolsep}{8pt} % Adjust column spacing
\small % Reduce font size slightly for better fit

\begin{tabular}{p{0.95\textwidth}} % Use 95% of text width to maintain margins
\toprule

\textbf{[QUESTION]}:
1. [OPTION 1] \quad | \quad 2. [OPTION 2] \quad | \quad 3. [OPTION 3] \quad | \quad 4. [OPTION 4] \\  

\textit{Please only provide the answer index.} \\  
\midrule
\textbf{Why did you choose ``I do not know''?} \\  
1. You were uncertain about the question and did not have enough knowledge to answer.  \\  
2. You thought the question was wrong and contained false information. \\  
\bottomrule
\end{tabular}

\caption{Evaluation prompts for multi-hop false premise reasoning where the models are supposed to answer with the index of the option. The top prompt assesses multiple-choice QA, where models may reject false premises by selecting ``I do not know''. The bottom prompt evaluates whether models correctly justify this choice.}
\label{tab:prompts}
\end{table*}

\iffalse
\begin{table*}[ht]
\small
\centering
\begin{tabular}{p{\textwidth}} 
\hline
[QUESTION]: 1. [FIRST OPTION] | 2. [SECOND OPTION] | 3. [THIRD OPTION] | 4. [FOURTH OPTION] \\

Please only provide the answer index.
\\
\hline
Why did you choose |I do not know?  \\
1. You were uncertain about the questions and did not have enough knowledge to answer.  \\
2. You thought the question was wrong and contained false information.
\\
\hline
\end{tabular}
\caption{Evaluation prompts.}
\label{tab:prompts}
\end{table*}
\fi

\subsection{Secondary False Information Verification}
To ensure the accuracy of falsified content, a second round of review was conducted. In this phase, a second reviewer independently examined the description field of each question against the corresponding Wikipedia page to verify the presence of false information. The reviewer categorized each question into one of three outcomes: ``There is false information'', ``There is no false information'', and ``I cannot tell based on the provided information''. \footnote{Table~\ref{tab:false_info_analysis_ann} in the appendix shows the distribution of the labels across categories.} Questions confirmed to contain false information were directly included in the final dataset. Those labeled with the second option were double-checked and falsehood was added where required, while those detected with the last option were both improved in terms of clarity and added with false information. In the latter two cases, after the necessary modifications, the reviewer provided feedback to ensure that the question contained clear false information. The dataset was finalized upon completion of this verification process.\footnote{Table~\ref{tab:annog} provides the annotation guidelines for this step.}
Table \ref{tab:MHFPQSEX} shows examples from our final resource from different types and countries.

\section{Evaluation Setup}
%\paragraph{Data} 

The {\texttt{MultiHoax}} dataset serves as the basis for our evaluation of multi-hop false premise reasoning through two primary tasks: (1) closed multiple-choice QA and (2) justification-based verification.
%\am{The \textbf{\texttt{MultiHoax}} dataset forms the foundation of our experiments, evaluating multi-hop false premise reasoning through several tasks: closed multiple-choice QA and justification. Moreover, we test models for the multiple-choice QA with JSON output format as well.
In the first task, models were presented with a question and four answer choices, including ``I do not know'', allowing them to reject the question if they identified a false premise. The prompt used for this step is shown in the first part of Table~\ref{tab:prompts}.

%In the first part, we presented the question with four options, including ``I do not know'', allowing for refusal to answer. The question and options along with the prompt shown in the first part of the Table \ref{tab:prompts} were provided to the LLMs. 

In the second task (justification), models that selected “I do not know” are prompted to explain their decision, as shown in the second part of Table~\ref{tab:prompts}. This step differentiates between cases where the model lacked knowledge and those where it explicitly identified a false premise. Only responses that both select ``I do not know'' in the first task and correctly justify it as a false premise in the second task can be considered successful detections of false premises. This justification step enhances the reliability of our evaluation, ensuring that refusal to answer stems from false premise detection rather than general uncertainty.

We additionally explore an alternative output format using structured JSON, where models are asked to produce both an answer and a textual explanation. While this format allows for more expressive reasoning, it can make it harder to unambiguously determine whether a model has identified a false premise, as rejection may be implied rather than explicit. By contrast, the multiple-choice setup includes an explicit “I do not know” option, enabling clearer detection of false-premise rejection. Despite the possibility that this option may simplify the task, it provides a more standardized and reliable evaluation framework. %Details of this setting and the corresponding prompt are presented in Section~\ref{sec:json} and Table~\ref{tab:prompt_json}.

%\am{Moreover, to allow explicit reasoning, we designed another output setting, where models were supposed to generate a JSON with an explanation and an answer field. The prompt used for this experiment is shown in Table \ref{tab:prompt_json}.}
%Secondly, if the model selected ``I do not know'', it was prompted to explain why, as shown in the second section of Table~\ref{tab:prompts}. This step helps assess whether the model lacked the knowledge to answer (was not really able to detect the falsehood) or detected falsehoods in the question. Serving as an additional measure of LLMs' ability to identify misinformation, this step further validates the results from the first task. Only the subset of the ``I do not know'' answers that later were accompanied by misinformation in the second task can be considered truly capable of detecting falsehoods.

\paragraph{Models}
We tested 6 different proprietary and open-source LLMs in our experiments. The models include Claude Sonnet 3.5\footnote{\url{https://www.anthropic.com/}}, Gemini-2.0-pro-exp \cite{team2023gemini}, GPT-4o \cite{hurst2024gpt}, Qwen2.5-7B-Instruct \cite{yang2024qwen2}, Llama-3.1-8B-Instruct \cite{meta2024llama}, and Deepseek-llm-7b-chat \cite{bi2024deepseek}. All experiments used a zero temperature setting to ensure deterministic responses, with all data collected in February 2025.

\section{Results}

\begin{table}[h]
\centering
\renewcommand{\arraystretch}{1.2}
\setlength{\tabcolsep}{8pt}
\footnotesize

\begin{tabular}{lcc}
\toprule
\textbf{Model} & \textbf{1st Task} & \textbf{2nd Task} \\
\midrule
Claude Sonnet 3.5        & \textbf{0.46} & 0.23 \\
Gemini-2.0-pro-exp       & 0.29          & \textbf{0.26} \\
GPT-4o-2024-11-20        & 0.23          & 0.25 \\
Qwen2.5-7B-Instruct      & 0.19          & 0.03 \\
Llama-3.1-8B-Instruct    & 0.13          & 0.01 \\
Deepseek-llm-7b-chat     & 0.05          & 0.06 \\
\bottomrule
\end{tabular}

\caption{Model performance on \texttt{MultiHoax}, evaluating multi-hop false premise reasoning in two tasks: (1) one-token multiple-choice QA, where models may reject false premises by selecting ``I do not know''; and (2) justification, where models must correctly explain that choice to confirm recognition of a false premise.}
\label{tab:models_acc}
\end{table}

\begin{table*}[htbp]
\centering
\footnotesize
\renewcommand{\arraystretch}{1.05}
\setlength{\tabcolsep}{5pt}

\begin{tabular}{lcccccc|c}
\toprule
\textbf{Category / Model} & \textbf{Claude} & \textbf{Gemini} & \textbf{GPT} & \textbf{Qwen} & \textbf{Llama} & \textbf{Deepseek} & \textbf{Avg} \\
\midrule
Science and Technology               & 0.458 & 0.329 & 0.286 & 0.215 & 0.143 & 0.430 & \textbf{0.310} \\
Entertainment                       & 0.429 & 0.286 & 0.172 & 0.129 & 0.115 & 0.000 & 0.189 \\
Education                           & 0.529 & 0.343 & 0.200 & 0.215 & 0.129 & 0.072 & 0.248 \\
Art and Literature                  & 0.458 & 0.258 & 0.172 & 0.072 & 0.086 & 0.043 & 0.182 \\
Food                                & 0.529 & 0.300 & 0.186 & 0.200 & 0.158 & 0.072 & 0.241 \\
Religion                            & 0.543 & 0.200 & 0.200 & 0.158 & 0.100 & 0.058 & 0.210 \\
Sports                              & 0.443 & 0.415 & 0.315 & 0.300 & 0.186 & 0.058 & 0.286 \\
Holiday, Celebrations, and Leisure  & 0.580 & 0.286 & 0.315 & 0.229 & 0.100 & 0.072 & 0.264 \\
Geography                           & 0.343 & 0.243 & 0.215 & 0.186 & 0.172 & 0.058 & 0.203 \\
History                             & 0.343 & 0.286 & 0.258 & 0.172 & 0.158 & 0.058 & 0.213 \\
\midrule
\textbf{Avg}                        & \textbf{0.466} & 0.295 & 0.232 & 0.188 & 0.135 & 0.092 & -- \\
\bottomrule
\end{tabular}

\caption{Accuracy of models across knowledge categories in \texttt{MultiHoax}. }
\label{tab:results_model_cat}
\end{table*}

\begin{table*}[htbp]
\centering
\footnotesize
\renewcommand{\arraystretch}{1.05}
\setlength{\tabcolsep}{5pt}

\begin{tabular}{lcccccc|c}
\toprule
\textbf{Country / Model} & \textbf{Claude} & \textbf{Gemini} & \textbf{GPT} & \textbf{Qwen} & \textbf{Llama} & \textbf{Deepseek} & \textbf{Avg} \\
\midrule
China    & 0.49 & 0.32 & 0.23 & 0.19 & 0.17 & 0.02 & 0.24 \\
France   & 0.45 & 0.29 & 0.23 & 0.21 & 0.15 & 0.06 & 0.23 \\
Germany  & 0.47 & 0.31 & 0.30 & 0.16 & 0.10 & 0.06 & 0.23 \\
Iran     & 0.47 & 0.34 & 0.25 & 0.19 & 0.13 & 0.05 & 0.24 \\
Italy    & 0.52 & 0.31 & 0.24 & 0.17 & 0.12 & 0.03 & 0.23 \\
UK       & 0.47 & 0.31 & 0.22 & 0.25 & 0.17 & 0.08 & \textbf{0.25} \\
USA      & 0.37 & 0.18 & 0.15 & 0.14 & 0.10 & 0.07 & 0.17 \\
\midrule
\textbf{Avg} & \textbf{0.46} & 0.29 & 0.23 & 0.19 & 0.13 & 0.06 & -- \\
\bottomrule
\end{tabular}

\caption{Accuracy of models across countries in \texttt{MultiHoax}. }
\label{tab:results_model_country}
\end{table*}

\begin{table*}[!tb]
\centering
\footnotesize
\renewcommand{\arraystretch}{1.05}
\setlength{\tabcolsep}{5pt}

\begin{tabular}{lcccccc|c}
\toprule
\textbf{Multi-hop Type} & \textbf{Claude} & \textbf{Gemini} & \textbf{GPT} & \textbf{Qwen} & \textbf{Llama} & \textbf{DeepSeek} & \textbf{Avg} \\
\midrule
Comparison     & 0.593 & 0.373 & 0.271 & 0.153 & 0.153 & 0.101 & \textbf{0.274} \\
Geographical   & 0.439 & 0.367 & 0.235 & 0.224 & 0.173 & 0.082 & 0.253 \\
Intersection   & 0.466 & 0.271 & 0.227 & 0.171 & 0.112 & 0.036 & 0.214 \\
Named          & 0.445 & 0.283 & 0.191 & 0.197 & 0.133 & 0.035 & 0.214 \\
Temporal       & 0.437 & 0.261 & 0.277 & 0.193 & 0.143 & 0.067 & 0.230 \\
\midrule
\textbf{Avg}   & \textbf{0.476} & 0.311 & 0.240 & 0.188 & 0.143 & 0.064 & -- \\
\bottomrule
\end{tabular}

\caption{Accuracy of models across different multi-hop reasoning types in \texttt{MultiHoax}. }
\label{tab:model_performance_mh_type}
\end{table*}

Table~\ref{tab:models_acc} presents model accuracy on the first group of tasks, which are the first two tasks.
The first task evaluates whether models correctly reject false premises by selecting ``I do not know''. While Claude, which was used during the question-generation phase, demonstrates higher accuracy compared to other models, the models generally struggle to recognize falsehoods in the first task, as they are unable to refuse to answer. Furthermore, although all models show poor performance on this task, open-source models generally exhibit lower accuracy than proprietary ones.

The second task analyzes whether models can justify their ``I do not know'' responses by correctly identifying the false premise. Here, Gemini slightly outperforms the generator model (Claude), but overall accuracy remains low across all models, indicating persistent difficulty in recognizing false premises.

%To further investigate why models choose to respond with ``I do not know,'' we introduce the second task of asking the models to justify their reason for answering with ``I do not know''. Specifically, they are asked whether they lacked the knowledge to answer or if the question contained incorrect information. In this case, Gemini outperforms the generator model. Nonetheless, all models continue to struggle with identifying false information, as indicated by their low accuracy.

Table~\ref{tab:results_model_cat} presents model performance across ten knowldege categories. Accuracy is highest in science and technology, while art and literature and entertainment rank lowest, reflecting variation in LLMs’ domain-specific factual competence.  Notably, proprietary models consistently outperform open-source models, though performance remains low across all categories. Interestingly, Claude's dominance is most pronounced in high-information domains like education and food, while other models exhibit irregular performance trends. These results highlight significant disparities in model strengths across different knowledge areas.

%Furthermore, Table~\ref{tab:results_model_cat} presents model performance across ten categories. Models perform best in science and technology, while art and literature show the lowest accuracy, suggesting different levels of LLMs' knowledge across categories. Overall, accuracy remains low across all categories, with proprietary models performing slightly better than open-source ones.

\begin{table*}[ht]
\centering
\renewcommand{\arraystretch}{1.2} % Adjust row height for better spacing
\setlength{\tabcolsep}{8pt} % Adjust column spacing
\small % Reduce font size slightly for better fit

\begin{tabular}{p{0.95\textwidth}} % Use 95% of text width to maintain margins
\toprule
\textbf{[QUESTION]}:\\
% 1. [FIRST OPTION] \quad | \quad 2. [SECOND OPTION] \quad | \quad 3. [THIRD OPTION] \quad | \quad 4. [FOURTH OPTION] \\  
Return the answer in a JSON format with a brief 'explanation' and an 'answer' field. \\
Here is an example output: \\
\{\\
"explanation": "The University of Tehran Press published the works of Hafez when it was founded."   \\
"answer": "Hafez"\\
\}
\\
ONLY RETURN THE JSON.
\\  
\bottomrule
\end{tabular}

\caption{Evaluation prompts for the JSON output structure format.}
\label{tab:prompt_json}
\end{table*}

Table~\ref{tab:results_model_country} analyzes accuracy across countries. The results indicate close accuracy levels, suggesting that models struggle to detect falsehoods regardless of the country. The lower accuracy in some countries can be a result of more challenging questions, which can suggest that the generator model was better able to design challenging questions for those countries.

Table~\ref{tab:model_performance_mh_type} presents model accuracy across multi-hop reasoning types. While comparison-based questions show slightly higher accuracy, the differences across reasoning types are modest, and overall performance remains low. This suggests that all multi-hop reasoning categories pose significant challenges to current models.\footnote{Tables~\ref{tab:model_performance_at} and ~\ref{tab:fptypeac} in the appendix provide model accuracy across different false premise and answer types.}

\begin{table}[h!]
\centering
\renewcommand{\arraystretch}{1.1}
\setlength{\tabcolsep}{10pt}
\footnotesize

\begin{tabular}{lc}
\toprule
\textbf{Model} & \textbf{Accuracy} \\
\midrule
Claude  & \textbf{0.46} \\
GPT     & 0.20 \\
Gemini  & 0.18 \\
Qwen    & 0.22 \\
\bottomrule
\end{tabular}

\caption{Accuracy of models when asked to provide structured JSON outputs with both answer and explanation fields.}
\label{tab:model_accuracy_json}
\end{table}

\iffalse
\begin{table}[h!]
\centering
\begin{tabular}{l|c}
\hline
\textbf{Model} & \textbf{Accuracy} \\
\hline
Claude & \textbf{0.46} \\
GPT & 0.20 \\
Gemini & 0.18 \\
Qwen & 0.22 \\
\hline
\end{tabular}
\caption{Model accuracy comparison where the models are asked to provide JSON outputs.}
\label{tab:model_accuracy_json}
\end{table}
\fi

{Finally, we evaluate models using a structured JSON output format that encourages explicit reasoning. This setting uses the prompt shown in Table~\ref{tab:prompt_json}.  Table \ref{tab:model_accuracy_json} presents the results of JSON output for the top-performing models from the previous task. The results indicate accuracy levels comparable to those observed in the multiple-choice setting. This suggests that models consistently struggle to detect embedded falsehoods in the \texttt{MultiHoax} dataset, regardless of output format or reasoning constraints.}

\section{Conclusion}
We introduce a novel class of multi-hop false-premise questions (MHFPQs), combining the complexities of multi-hop reasoning and false-premise detection. To support this research, we present \texttt{MultiHoax}, the first FPQ resource that spans multiple countries and knowledge domains, enabling evaluation of LLMs' ability to navigate multi-step falsehoods in diverse contexts.
Our dataset provides a comprehensive evaluation framework, spanning seven countries and ten knowledge categories, allowing for a detailed analysis of how LLMs handle false premises across diverse topics and regions. Unlike prior FPQ datasets, MHFPQs require deeper reasoning, as falsehoods are not immediately apparent but emerge through multi-step inference.
\texttt{MultiHoax} establishes a novel evaluation setting at the intersection of multi-hop reasoning and false-premise detection, revealing persistent limitations in current LLMs. By targeting assumption-level failures across diverse domains and regions, it provides a rigorous benchmark for advancing models that can reason robustly under implicit factual errors.

%We introduce a novel class of MHFPQs, exploring their various types through different combinations of MHQs and FPQs. To support this research, we present \texttt{MultiHoax}, the first cross-regional FPQs resource, which extends beyond previously known FPQs to focus on MHFPQs—a more complex and challenging question type.
%MHFPQs assess LLMs' ability to detect falsehoods within questions and refrain from answering misleading queries. These falsehoods are not immediately obvious and require multi-step reasoning to identify.

%Our dataset includes a comprehensive evaluation set of MHFPQs spanning seven countries and ten categories, enabling scholars to analyze LLMs' effectiveness in detecting nuanced misinformation across different regions and topics.

%We anticipate that this dataset will benefit a wide range of research areas, including cultural and regional studies, QA research, and misinformation detection, providing a valuable benchmark for evaluating LLMs' ability to navigate complex falsehoods in questions.

\section*{Limitations}
Our dataset has some limitations. While we have aimed to include a diverse range of countries with varying levels of resource availability, there are still opportunities to incorporate additional countries from other regions worldwide.

Second, our category set, which consists of 10 categories, could be expanded as scholars explore knowledge across various areas. While we have focused on a set of categories suitable for factual, objective questions, other potential categories could be included. Additionally, subjective questions, such as ``Who first imported the most popular type of ingredient to China?'' could be considered. This would be an MHQ, as it requires identifying both the most popular ingredient and the first person to import it. However, determining the most popular ingredient is subjective, and the question becomes MHFP if the ingredient was never imported to China. While subjective questions are feasible, reviewing them differs significantly from the process for factual objective questions.

Furthermore, while we include questions associated with a diverse set of countries, we do not have the translation of these questions in the local languages of these countries, except from the United States and the United Kingdom. Future research can be collecting questions in the local language of such countries or translating ours to those languages.

Finally, while our current resource presents a significant challenge to different LLMs, and even the best models struggle with our tasks, the rapid progress of LLMs may make our dataset less difficult over time. As models improve, we may need to update our resources to introduce more challenging tasks that better test LLMs' reasoning abilities.

\section*{Ethical Considerations}
We aim to provide a set of factually incorrect questions requiring multiple reasoning steps to challenge various models' ability to detect falsehoods.  

To compile our dataset, we utilized LLMs to generate potential questions, which greatly facilitated the process. However, this approach may introduce biases, as LLMs are more knowledgeable about certain countries than others. For instance, the USA's lower accuracy could be attributed to LLMs having more in-depth knowledge of U.S. facts, allowing them to craft more challenging questions. Although we used Wikipedia documents to mitigate this bias, it cannot be entirely eliminated.  
In future iterations, we plan to diversify the question types, incorporating topics focused on values and norms rather than just factual knowledge, and we aim to minimize reliance on LLMs for question generation.

\bibliography{custom}

\appendix
\section*{Appendix A: Prompts}
\label{sec:appendixa}

\begin{table*}
\centering
\begin{tabular}{p{0.95\textwidth}}
\hline

We are developing a dataset of factual questions and aiming to collect questions across various categories and countries.  
Currently, we are identifying relevant important Wikipedia pages for each category and country.  
Please gather Wikipedia pages related to \textbf{[CATEGORY]} for the following countries:
\begin{itemize}
    \item China
    \item France
    \item Germany
    \item Italy
    \item United States
    \item United Kingdom
    \item Iran
\end{itemize}

Please provide \textbf{15} distinct Wikipedia pages related to \textbf{[CATEGORY]} for the mentioned countries, based on the following category definition below.
\\
\texttt{[DEFINITION OF THE CATEGORY]}

ONLY PROVIDE THE PAGE NAMES LINKED TO THE PAGE WITHOUT ANY EXPLANATION.

\\

\hline
\end{tabular}
\caption{The prompt for searching for related documents using ChatGPT4o.}
\label{tab:doc_gpt_ret}
\end{table*}

In this section, we provide the different prompts that we leveraged to collect the Wikipedia pages, and also prompts to generate the questions.

\subsection*{Wikipedia Document Retrieval Prompt}
In this section, we first define each category. Then, we used them in order to search for the relevant documents of each category using ChatGPT-4o's search using the prompt in Table \ref{tab:doc_gpt_ret}.
\paragraph{Food}
\begin{itemize}
    \item \textbf{Cuisine}: Signature dishes, cooking styles, traditional meals.
    \item \textbf{Ingredients}: Locally grown spices, crops, and special ingredients.
    \item \textbf{Drinks}: Popular beverages, traditional teas, or alcoholic drinks.
\end{itemize}

\paragraph{Sports}
\begin{itemize}
    \item \textbf{National and Popular Sports}: Widely played or watched sports in the country and Official sports of a country.
    \item \textbf{Athletes}: Famous sportspeople or Olympic medalists.
    \item \textbf{Tournaments and Sports Venues}: Major leagues, championships, or cups, as well as iconic stadiums, arenas, or tracks.
\end{itemize}

\paragraph{Education}
\begin{itemize}
    \item \textbf{Education System AND Literacy}: Structure (primary, secondary, higher education) AND Efforts to promote literacy or improve access to education.
    \item \textbf{Schools and Universities AND Curriculum}: Prestigious or historic institutions AND Subjects emphasized or unique courses.
    \item \textbf{Famous Educators}: Scholars, reformers, or pioneers in education.
\end{itemize}

\paragraph{Holidays/Celebrations/Leisure}
\begin{itemize}
    \item \textbf{National Holidays}: Independence days, constitution days, or memorials.
    \item \textbf{Festivals}: Cultural, religious, or seasonal festivals.
    \item \textbf{Others}: Other topics related to Holidays/Celebrations/Leisure.
\end{itemize}

\paragraph{History}
\begin{itemize}
    \item \textbf{Historical Figures}: Leaders, revolutionaries, empires and kingdoms, or intellectuals.
    \item \textbf{Important Events}: Battles, treaties, or turning points in history.
    \item \textbf{Landmarks}: Historical monuments or UNESCO heritage sites.
\end{itemize}

\paragraph{Geography}
\begin{itemize}
    \item \textbf{Natural Features AND Resources}: Mountains, rivers, lakes, and deserts AND Natural resources, agriculture, or energy production.
    \item \textbf{Cities AND Regions}: Capitals, major cities, or urban landmarks AND Administrative divisions or cultural regions.
    \item \textbf{Geopolitics}: Borders, neighbors, or disputed territories.
\end{itemize}

\paragraph{Science and Technology}
\begin{itemize}
    \item \textbf{Scientists}: Modern renowned scientists.
    \item \textbf{Engineering}: Famous modern constructions, bridges, or technology.
    \item \textbf{Others}: Other related topics to Science and Technology like Medical Breakthroughs, Research Centers, Computing Pioneers, Green Technology, Digital Platforms, and Communications.
\end{itemize}

\paragraph{Arts and Literature}
\begin{itemize}
    \item \textbf{Writers}: Prominent authors, poets, or playwrights.
    \item \textbf{Books}: National epics, famous novels, or historical documents.
    \item \textbf{Artists}: Prominent artists.
\end{itemize}

\paragraph{Religion}
\begin{itemize}
    \item \textbf{Religions and Religious Practices}: Popular religions and Worship styles or Religious rituals.
    \item \textbf{Holy Sites}: Temples, churches, mosques, or pilgrimage locations.
    \item \textbf{Others}: Religious Leaders, Religious Festivals, or Sacred Texts.
\end{itemize}

\paragraph{Entertainment}
\begin{itemize}
    \item \textbf{Cinema and TV}: National cinema, famous directors, popular movies, or actors.
    \item \textbf{Music}: Traditional music styles, Musicians, or iconic bands.
    \item \textbf{Others}: Other topics related to entertainment like theater, gaming, festivals related to entertainment, or media.
\end{itemize}

These category definitions were used in the prompt in the Table \ref{tab:doc_gpt_ret} to get 15 related pages for each country.

\subsection*{Fact Extraction Prompt}
\label{sec:fep}
To extract facts from each document, we leveraged a simple brief prompt. In the fact extraction prompt, we define the number of facts to be extracted, the name of the document, and the document content. The prompt is provided in the Table \ref{tab:prompt_fact_ex}.

\begin{table}
\centering
\begin{tabular}{c}
\hline

Extract 15 facts from the document. \\
Document Content: \textbf{DOCUMENT CONTENT}
\\

\hline
\end{tabular}
\caption{Fact extraction prompt.}
\label{tab:prompt_fact_ex}
\end{table}

\begin{table}
    \centering
    \begin{tabular}{cc}
        \hline
        Type & Percentage (\%)\\ 
        \hline
        Intersection & 36 \\ 
        Named-entity & 25 \\ 
        Temporal-entity & 17 \\ 
        Geographical-entity & 14 \\ 
        Comparison & 8 \\ 
        \hline
    \end{tabular}
    \caption{Distribution of MH types in the dataset.}
    \label{tab:fptypescount}
\end{table}

\begin{table}
    \centering
    \small
    \begin{tabular}{cc}
        \hline
        Type & Percentage (\%) \\ 
        \hline
        Person & 44 \\ 
        Group/Org & 13 \\ 
        Location & 8 \\ 
        Number & 7 \\ 
        Language or Nationality or Country & 6 \\ 
        Date & 6 \\ 
        Common noun & 5 \\ 
        Food & 4 \\
        Artwork & 3 \\ 
        % Adjective & 1 \\ Added to other proper noun 
        Event & 2 \\ 
        Other proper noun & 2 \\ 
        \hline
    \end{tabular}
    \caption{Distribution of Answer types in the dataset.}
    \label{tab:optcount}
\end{table}

\subsection*{Question Generation Prompt}
\begin{table*}
\centering
\begin{tabular}{p{\textwidth}}
\hline

We are developing a dataset of counterfactual multi-hop reasoning false-premise questions (MHFPQs).
Multi-hop reasoning questions require retrieving and connecting multiple pieces of information across two or more logical steps to derive the final answer.
We have seen what a multi-hop question is.

Now, let's focus on creating questions with false premises—where in the question there exists a false assumption that makes the question wrong and impossible to answer correctly. 
MHFPQs have similar types to Multi-hop but with false premises.

Here are some examples: Note that these are only examples, DO NOT include them in your answers.
\\

\hline
\end{tabular}
\caption{The first part of the generation prompt.}
\label{tab:gen_prompt_1}
\end{table*}

For question generation, we designed two distinct prompts corresponding to the two major types of MHQs: Entity-based and Intersection \& Comparison Combination. Each prompt consists of three parts: an introduction to the task, an explanation of the specific MHQ type, and a set of generation rules.
The first section, shown in the Table \ref{tab:gen_prompt_1}, and the last section, shown in the Table \ref{tab:gen_prompt_3} are the same in both generation prompts, but the second part, the definition of the specific MHQ types, is different.

The tables \ref{tab:gen_prompt_2_1} and \ref{tab:gen_prompt_2_2} show the two different second sections.

\begin{table*}
\centering
\begin{tabular}{p{\textwidth}}
\hline
\vspace{1mm}
\begin{minipage}{\textwidth}
\begin{itemize}
    \item \textbf{Temporal Multi-Hop Reasoning} \\
    \textit{Example}: Who was the president of Iran in the year in which Ehsan Rouzbahani won the Olympics Bronze medal in Tokyo?
    
    Description: Ehsan Rouzbahani did not compete in the Tokyo Olympics, making it impossible for him to win a (bronze) medal.
    
    Reasoning Steps:
    \begin{enumerate}
        \item The year when Ehsan Rouzbahani won the Olympic bronze medal in Tokyo.
        \item The president of Iran at that time.
    \end{enumerate}
    
    \item \textbf{Geographical Multi-Hop Reasoning} \\
    \textit{Example}: Which football team with the most championships in the territory Alexander the Great conquered before turning 18?
    
    Description: Alexander the Great did not conquer any territory before turning 18.
    
    Reasoning Steps: 
    \begin{enumerate}
        \item The territory Alexander the Great conquered before turning 18.
        \item Football team with the most championships in that territory.
    \end{enumerate}
    
    \item \textbf{Named Entity Multi-Hop Reasoning} \\
    \textit{Example}: What is the camera brand used by Spielberg when filming his Academy Award-winning student film at USC? 
    
    Description: Spielberg never attended USC and didn't win an Academy Award as a student.
    
    Reasoning Steps: 
    \begin{enumerate}
        \item The Spielberg's Academy Award-winning student film at USC.
        \item The camera brand used for that film.
    \end{enumerate}
\end{itemize}
\end{minipage} \\\\
\hline
\end{tabular}
\caption{The second part of the generation prompt for the first group.}
\label{tab:gen_prompt_2_1}
\end{table*}

\begin{table*}
\centering
\begin{tabular}{p{\textwidth}}
\hline
\vspace{1mm}
\begin{minipage}{\textwidth}
\begin{itemize}
    \item \textbf{Intersection-Type Multi-Hop Reasoning} \\
    \textit{Example}: Which architect both designed the golden-domed Old Basilica and incorporated Aztec symbols in its facade in 1695?
    
    Description: The Old Basilica did not have Aztec symbols. It only had yellow and blue Talavera mosaics.
    
    Reasoning Steps:
    \begin{enumerate}
        \item The architect who designed the golden-domed Old Basilica.
        \item The architect who incorporated Aztec symbols in the Old Basilica's facade in 1695.
    \end{enumerate}
    
    \item \textbf{Comparison-Type Multi-Hop Reasoning} \\
    \textit{Example}: Which of the Chinese and the Germans first invented sauerkraut in the 18th century?
    
    Description: Sauerkraut was not invented in the 18th century, and it existed much earlier.
    
    Reasoning Steps: 
    \begin{enumerate}
        \item Did the Chinese invented sauerkraut in the 18th century?
        \item Did Germans invented sauerkraut in the 18th century?
    \end{enumerate}
    
    Note: In Comparison type, multi-hop reasoning questions, none of the entities satisfies the condition.
\end{itemize}
\end{minipage} \\
\hline
\end{tabular}
\caption{The second part of the generation prompt for the second group.}
\label{tab:gen_prompt_2_2}
\end{table*}

\begin{table*}
\centering
\begin{tabular}{p{\textwidth}} % Use \textwidth to span the entire page width
\hline
Your task is to extract false premise multi-hop questions FROM THE FACTS PROVIDED. Here are the instructions:

\begin{itemize}
    \item The structure of the question should be like the given structures, but the content can be different.
    \item False premises are implicitly embedded within the questions. Also, false premises must not be obvious.
    \item Provide 3 relevant, engaging, and realistic options.
    \item Questions should be based on the provided FACTS from the specific country.
    \item Focus is exclusively on verifiable factual claims, avoiding cultural norms or subjective topics.
    \item For each question, a clear explanation must be provided:
    \item Identifying the false premise.
    \item Clarifying the actual truth.
    \item If it is not possible to design questions from all the types, you can only focus on most probable ones.
\end{itemize}

Return each question in the following format:

<false\_premise\_multi\_hop\_question> | <first\_option> | <second\_option> | <third\_option> | <description> | <reasoning\_steps> | <type\_of\_multi\_hop>\\
\hline
\end{tabular}
\caption{The third part of the generation prompt.}
\label{tab:gen_prompt_3}
\end{table*}

% \newpage  % or \clearpage
\section*{Appendix B: Annotation Guideline}
\label{sec:appendixb}
\subsection*{Annotation Guideline}
The annotation guideline, shown in the Table \ref{tab:annog} was used to familiarize false information reviewers with the dataset structure and their tasks. The guideline first defines what an FPQ is, then explains how questions are structured in our dataset, and finally outlines the reviewing task.

\begin{table*}[ht]
\centering
\begin{tabular}{p{\textwidth}} % Use \textwidth to span the entire page width
\hline
\textbf{ False-premise Questions}: We have a set of false-premise questions. A false premise question is a question with at least one piece of false information. A simple example of false-premise questions can be “How many eyes does the sun have?” Such simple questions include false information that is easily detectable by humans. More challenging false-premise questions, which are the target of our experiment, have false pieces of information that are difficult for non-expert humans to detect.\\
\\
\textbf{Question}: During which time, 1985 to 1990 or 1995 to 2010, did CERN's affiliates win more Nobel prizes in physics?
\begin{enumerate}
    \item 1985 to 1990
    \item 1995 to 2010
    \item In both durations, CERN's affiliates won only 1 Nobel prize
    \item I don’t know
\end{enumerate}
\\
\textbf{Explanation}: In none of the durations, CERN's affiliates won a Nobel prize. 1984, 1992, and 2013 are the years when CERN's affiliates won the award.

As you can see, such false information types are not detectable unless the person knows about the history of the mentioned institute, which is not the case for non-experts. 
\\ 
\textbf{Format of Data}: In the dataset, there are a number of fields, like the following example.
\\ 
\textbf{Question}: What is the name of the new Humanistic Buddhist organization that was established in Beijing in the 2000s to promote the revival of Vajrayana Buddhism in China?
Options: 
\begin{enumerate}
    \item Cíjì
    \item Huácáng Zōngmén
    \item Zhēnfó Zōng
    \item I don't know
\end{enumerate}
\textbf{Description}: The question contains a false premise that a new Humanistic Buddhist organization was established in Beijing in the 2000s to promote the revival of Vajrayana Buddhism. According to the facts, the Humanistic Buddhist movement in China is associated with organizations like Cíjì, which has been working in mainland China since 1991, not a new organization focused on Vajrayana Buddhism.
\textbf{Wikipedia}: https://en.wikipedia.org/wiki/Buddhism\_in\_China
\\ 
\textbf{Your task}: You are supposed to read the question and options, and then check the provided description, which is the description of why the question includes false information.
Afterward, you are supposed to label these questions after checking if there is any false information in them or not. “\textit{There is false information}”,  “\textit{There is no false information}”, and “\textit{I cannot tell based on the provided information}” are the possible options. You need to visit the Wikipedia page related to each question and check false information based on that \textbf{Wikipedia} page. 
\begin{itemize}
    \item If you choose “\textit{There is no false information}”, then you are supposed to explain why you have chosen this. For example, in the above case, if you choose “There is no false information”, then an example explanation can be “CERN’s affiliates won the Nobel prize in 1986 making the question true”.
    \item If you choose “\textit{I cannot tell based on the provided information}”, you must also explain the ambiguity or the problem you have in verifying the question in the explain column.
    \item If you choose “\textit{There is false information}”, then there is no need to explain.
\end{itemize}
Keep the explanation clear, simple, and concise.
\\
\hline
\end{tabular}
\caption{Reviewing guidelines.}
\label{tab:annog}
\end{table*}

\subsection*{Annotators details}
Following the initial round of annotation by the authors, the dataset was divided into three parts for the second review phase. We then recruited three university student annotators—one female and two male—each paid £12.21 per hour for ten hours of work.

% \newpage  % or \clearpage
\section*{Appendix C: Statistics}
\label{sec:appendixc}
In this section, we provide further statistical information regarding the dataset.

Table \ref{tab:fptypescount} provides the distribution of the various types of MHQs.The majority of the resource consists of bridge entity-based questions, while the intersection type is largely included.

\begin{table*}
    \centering
    \begin{tabular}{ccc}
        \hline
        Type & Description & Percentage (\%) \\ 
        \hline
        Event      & The event didn’t happen in history. & 49  \\
        Property   & The entity does not have the property. & 36 \\
        Scope      & A fact is not valid in the scope. & 13 \\
        Entity     & The entity cannot exist. & 2 \\
        % Index      & The specified index is out of an entity list. & & 1  \\ Added to Event
        \hline
    \end{tabular}
    \caption{Distribution of FP types in the dataset.}
    \label{tab:mhtypecount}
\end{table*}

Table \ref{tab:mhtypecount} shows the distribution of the various types of false information included in the questions. We include four types in our resources. The description of types is also provided in Table \ref{tab:mhtypecount}.

Moreover, Table \ref{tab:optcount} provides the details of answer type occurrence over the resource, the person type as the major type.

Regarding the annotation details, Table~\ref{tab:false_info_analysis_ann} contains the number of times the second reviewer chose each possible label for the questions.

\begin{table*}
\centering
\begin{tabular}{l|ccc} % Use \textwidth to span the entire page width
\hline
\textbf{File} & \textbf{False Information} & \textbf{Cannot Tell} & \textbf{No False Information} \\
\hline
Science and technology & 49 & 12 & 9 \\
Entertainment & 66 & 3 & 1 \\
Education & 65 & 5 & 0 \\
Art and Literature & 59 & 11 & 0 \\
Food & 62 & 6 & 2 \\
Religion & 59 & 6 & 5 \\
Sports & 56 & 8 & 6 \\
Holiday, Celebrations, and Leisure & 52 & 4 & 14 \\
Geography & 51 & 3 & 16 \\
History & 57 & 0 & 13 \\
\hline
Sum & 576 & 57 & 67 \\
\hline
\end{tabular}
\caption{Analysis of false information inclusion across different categories based on the second phase review.}
\label{tab:false_info_analysis_ann}
\end{table*}

The table \ref{tab:model_performance_at} provides the detailed accuracy of the models across different types of answers. As shown, event, location, and number have the lowest accuracies, suggesting that the models find these types more challenging than the rest while they generally show poor performance.

Tables \ref{tab:model_performance_mh_type} and \ref{tab:fptypeac} also show the detailed accuracies for different types of MHQs and FPQs.

\begin{table*}[h]
\centering
\footnotesize
\renewcommand{\arraystretch}{1.1}
\setlength{\tabcolsep}{5pt}

\begin{tabular}{lcccccc|c}
\toprule
\textbf{Answer Type} & \textbf{Claude} & \textbf{Gemini} & \textbf{GPT} & \textbf{Qwen} & \textbf{Llama} & \textbf{DeepSeek} & \textbf{Avg} \\
\midrule
Artwork                         & 0.480 & 0.360 & 0.280 & 0.160 & 0.080 & 0.040 & 0.233 \\
Common Noun                     & 0.459 & 0.351 & 0.243 & 0.243 & 0.162 & 0.162 & 0.270 \\
Date or Time Period             & 0.600 & 0.450 & 0.375 & 0.275 & 0.200 & 0.025 & 0.321 \\
Event                           & 0.429 & 0.143 & 0.214 & 0.143 & 0.071 & 0.000 & 0.167 \\
Food                            & 0.577 & 0.308 & 0.192 & 0.192 & 0.038 & 0.077 & 0.231 \\
Group or Org                    & 0.494 & 0.205 & 0.241 & 0.181 & 0.108 & 0.060 & 0.215 \\
Language/Nationality/Country    & 0.595 & 0.238 & 0.238 & 0.167 & 0.167 & 0.095 & 0.250 \\
Location                        & 0.339 & 0.271 & 0.186 & 0.136 & 0.136 & 0.034 & 0.184 \\
Number                          & 0.333 & 0.333 & 0.208 & 0.146 & 0.167 & 0.000 & 0.198 \\
Other Proper Noun               & 0.417 & 0.250 & 0.167 & 0.167 & 0.167 & 0.167 & 0.223 \\
Person                          & 0.453 & 0.296 & 0.219 & 0.193 & 0.135 & 0.045 & 0.223 \\
\midrule
\textbf{Avg}                    & {0.477} & 0.278 & 0.226 & 0.185 & 0.125 & 0.066 & 0.226 \\
\bottomrule
\end{tabular}

\caption{Accuracy of models across different answer types in \texttt{MultiHoax}.}
\label{tab:model_performance_at}
\end{table*}

\iffalse
\begin{table*}[h]
    \centering
    \begin{tabular}{l|cccccc|c}
    \hline
        \textbf{Answer Type} & \textbf{Claude} & \textbf{Gemini} & \textbf{GPT} & \textbf{Qwen} & \textbf{Llama} & \textbf{DeepSeek} & \textbf{Avg} \\
        \hline
        Artwork & 0.48 & 0.36 & 0.28 & 0.16 & 0.08 & 0.04 & 0.233 \\
        Common Noun & 0.459 & 0.351 & 0.243 & 0.243 & 0.162 & 0.162 & 0.270 \\
        Date or Time Period & 0.6 & 0.45 & 0.375 & 0.275 & 0.2 & 0.025 & 0.321 \\
        Event & 0.429 & 0.143 & 0.214 & 0.143 & 0.071 & 0.0 & 0.167 \\
        Food & 0.577 & 0.308 & 0.192 & 0.192 & 0.038 & 0.077 & 0.231 \\
        Group or Org & 0.494 & 0.205 & 0.241 & 0.181 & 0.108 & 0.060 & 0.215 \\
        Language/Nationality/Country & 0.595 & 0.238 & 0.238 & 0.167 & 0.167 & 0.095 & 0.250 \\
        Location & 0.339 & 0.271 & 0.186 & 0.136 & 0.136 & 0.034 & 0.184 \\
        Number & 0.333 & 0.333 & 0.208 & 0.146 & 0.167 & 0.0 & 0.198 \\
        Other Proper Noun & 0.417 & 0.25 & 0.167 & 0.167 & 0.167 & 0.167 & 0.223 \\
        Person & 0.453 & 0.296 & 0.219 & 0.193 & 0.135 & 0.045 & 0.223 \\
        \hline
        Avg & 0.477 & 0.278 & 0.226 & 0.185 & 0.125 & 0.066 & 0.226 \\
        \hline
    \end{tabular}
    \caption{Accuracy of the models across different answer types.}
    \label{tab:model_performance_at}
\end{table*}
\fi

\begin{table*}[h]
\centering
\footnotesize
\renewcommand{\arraystretch}{1.1}
\setlength{\tabcolsep}{6pt}

\begin{tabular}{lcccccc|c}
\toprule
\textbf{FP Type} & \textbf{Claude} & \textbf{Gemini-2.0} & \textbf{GPT-4} & \textbf{Qwen2.5} & \textbf{Llama-3.1} & \textbf{DeepSeek-7B} & \textbf{Avg} \\
\midrule
Entity   & 0.778 & 0.510 & 0.333 & 0.278 & 0.278 & 0.111 & 0.381 \\
Event    & 0.451 & 0.309 & 0.209 & 0.197 & 0.151 & 0.062 & 0.230 \\
Property & 0.478 & 0.251 & 0.247 & 0.183 & 0.112 & 0.048 & 0.220 \\
Scope    & 0.400 & 0.311 & 0.244 & 0.144 & 0.111 & 0.022 & 0.205 \\
\midrule
\textbf{Avg} & \textbf{0.527} & 0.345 & 0.258 & 0.201 & 0.163 & 0.061 & 0.259 \\
\bottomrule
\end{tabular}

\caption{Accuracy of models across different false premise (FP) types in \texttt{MultiHoax}.}
\label{tab:fptypeac}
\end{table*}

\iffalse
\begin{table*}[h]
\centering
\begin{tabular}{l|cccccc|c}
\hline
\textbf{FP Type} & \textbf{Claude} & \textbf{Gemini-2.0} & \textbf{GPT-4} & \textbf{Qwen2.5} & \textbf{Llama-3.1} & \textbf{DeepSeek-7B} & \textbf{Avg} \\
\hline
Entity & 0.778 & 0.510 & 0.333 & 0.278 & 0.278 & 0.111 & 0.381 \\
Event & 0.451 & 0.309 & 0.209 & 0.197 & 0.151 & 0.062 & 0.230 \\
Property & 0.478 & 0.251 & 0.247 & 0.183 & 0.112 & 0.048 & 0.220 \\
Scope & 0.400 & 0.311 & 0.244 & 0.144 & 0.111 & 0.022 & 0.205 \\
\hline
\textbf{Avg } & 0.527 & 0.345 & 0.258 & 0.201 & 0.163 & 0.061 & 0.259 \\
\hline
\end{tabular}
\caption{Accuracy of the models across different FP types.}
\label{tab:fptypeac}
\end{table*}
\fi

\end{document}